\documentclass{article}

\usepackage{microtype}
\usepackage{graphicx}
\usepackage{subfigure}
\usepackage{booktabs} % for professional tables

\usepackage{hyperref}

\usepackage[utf8]{inputenc}
\usepackage{xcolor}
\usepackage{algorithm}
\usepackage{algorithmic}
\usepackage{amsthm}
\usepackage{graphicx}
\usepackage[multiple]{footmisc}

\newtheorem{theorem}{Theorem}[section]
\newtheorem{lemma}[theorem]{Lemma}
\newtheorem{remark}[theorem]{Remark}
\newtheorem{corollary}[theorem]{Corollary}

\newtheorem{example}[theorem]{Example}

\newcommand{\esm}[1]{\ensuremath{#1}}

\newcommand{\ms}[1]{\esm{\mathsf{#1}}}

\newcommand\reals{\ms{R}}

\newcommand\sparam{\alpha}

\DeclareTextFontCommand{\emph}{\bfseries}
\begin{document}

%\twocolumn[
%\icmltitle{The Many Shapley Values for Model Explanation}
\title{The Many Shapley Values for Model Explanation}

% It is OKAY to include author information, even for blind
% submissions: the style file will automatically remove it for you
% unless you've provided the [accepted] option to the icml2020
% package.

% List of affiliations: The first argument should be a (short)
% identifier you will use later to specify author affiliations
% Academic affiliations should list Department, University, City, Region, Country
% Industry affiliations should list Company, City, Region, Country

% You can specify symbols, otherwise they are numbered in order.
% Ideally, you should not use this facility. Affiliations will be numbered
% in order of appearance and this is the preferred way.
%\icmlsetsymbol{equal}{*}

%\begin{icmlauthorlist}

%\end{icmlauthorlist}

\author{
  Mukund Sundararajan\\
  \texttt{mukunds@google.com}
  \and
  Amir Najmi\\
  \texttt{amir@google.com}
}

%\icmlaffiliation{goo}{Google LLC}

%\icmlcorrespondingauthor{Cieua Vvvvv}{c.vvvvv@googol.com}
%\icmlcorrespondingauthor{Eee Pppp}{ep@eden.co.uk}

% You may provide any keywords that you
% find helpful for describing your paper; these are used to populate
% the "keywords" metadata in the PDF but will not be shown in the document
%\icmlkeywords{Machine Learning, ICML}

\vskip 0.3in
%]

% this must go after the closing bracket ] following \twocolumn[ ...

% This command actually creates the footnote in the first column
% listing the affiliations and the copyright notice.
% The command takes one argument, which is text to display at the start of the footnote.
% The \icmlEqualContribution command is standard text for equal contribution.
% Remove it (just {}) if you do not need this facility.

%\printAffiliationsAndNotice{}  % leave blank if no need to mention equal contribution
%\printAffiliationsAndNotice{\icmlEqualContribution} % otherwise use the standard text.
\maketitle

\begin{abstract}

The Shapley value has become a popular method to attribute the prediction of a machine-learning model on an input to its base features. The use of the Shapley value is justified by citing ~\cite{Shapley53} showing that it is the \emph{unique} method that satisfies certain good properties (\emph{axioms}). 

There are, however, a multiplicity of ways in which the Shapley value is operationalized in the attribution problem. These differ in how they reference the model, the training data, and the explanation context. These give very different results, rendering the uniqueness result meaningless. Furthermore, we find that previously proposed approaches can produce counterintuitive attributions in theory and in practice---for instance, they can assign non-zero attributions to features that are not even referenced by the model. 

In this paper, we use the axiomatic approach to study the differences between some of the many operationalizations of the Shapley value for attribution, and propose a technique called Baseline Shapley (BShap) that is backed by a proper uniqueness result. We also contrast BShap with Integrated Gradients, another extension of Shapley value to the continuous setting.
 
\end{abstract}

\section{Motivation and Related Work}
\label{sec:intro}

We discuss the \emph{attribution problem}, i.e., the problem of distributing the prediction score of a model for a specific input to its base features (cf.~\cite{Lime, Lundberg2017AUA, STY17}); the attribution to a base feature can be interpreted as the importance of the feature to the prediction. For instance, when attribution is applied to a model that makes loan decisions, the attributions tell you how influential a feature was to the loan decision for a specific loan applicant. Attributions thus have explanatory value.

One of the leading approaches to attribution is based on the Shapley value~\cite{Shapley53}, a construct from cooperative game theory. In cooperative game theory, a group of players come together to consume a service, and this incurs some cost. The Shapley value distributes this cost among the players. There is a correspondence between  cost-sharing and the attribution problem: The cost function is analogous to the model, the players to base features, and the cost-shares to the attributions. 

The Shapley value is known to be the unique method that satisfies certain properties (see Section~\ref{sec:shapley} for more details). The desirability of these properties, and the uniqueness result make a strong case for using the Shapley value. Unfortunately, despite the uniqueness result, there are a multiplicity of Shapley values that differ in how they refer to the model, the training data, and the explanation context. Here is a chronological sampling of the literature: 

\begin{enumerate}
    \item There is literature (cf. ~\cite{LMG,Ulrike}) that uses the Shapley value to attribute the goodness of fit ($R^2$) of a linear regression model to its features by retraining the model on different feature subsets. 
    \item  \cite{Owen1,Owen2} apply the Shapley value to study the importance of a feature to a given function, by using it to identify the "variance explained" by the feature; no retraining involved .
    \item  \cite{strumbelj09, Strumbelj} use the Shapley value to solve the attribution problem, i.e., feature importance for a specific prediction. The first paper applies the Shapley value by retraining the model on every possible subset of the features. The second paper applies the Shapley value to the conditional expectation of a specific model (no retraining) (see Section~\ref{sec:conditional-expecations} for a formal definition of the conditional expectation approach). They assume that features are distributed uniformly and independently.
    \item \cite{Datta} applies the Shapley value to the conditional expectations of the model's function with a contrived distribution that is the product of the marginals of the underlying feature distribution.  
    \item \cite{Lundberg2017AUA} also investigates the Shapley value with conditional expectations; it constructs various approximations that make assumptions about either the function, or the distribution, and applies it compositionally on modules of a deep network.
    \item \cite{Lundberg18} computes the Shapley value with conditional expectations efficiently for trees; however, it is not very clear about its assumptions on the feature distribution \footnote{In an email exchange, Scott Lundberg clarified that the implicit assumption is that the features are distributed according to "the distribution generated by the tree".}. 
    \item \cite{Aas} generalizes one of the approaches in~\cite{Lundberg2017AUA} to the case when the distributions are not independent, either by assuming that the features are generated by a mixture of Gaussians or by a non-parametric, heuristic approach that applies the Mahalanobis distance to the empirical distribution.
    \item Unlike the methods above that either delete or marginalize over a feature, \cite{Multilinear, STY17, ADS} apply the Shapley value, by using a different approach to `turn features off'. This approach takes an auxiliary input called a baseline, and switches the explicand's feature value to the value of the feature in the baseline (see Section~\ref{sec:baseline} for details).
    \item \cite{STY17} proposes a technique called Integrated Gradients, that is based on the  Aumann-Shapley~\cite{AS74} cost-sharing technique. Aumann-Shapley is one of the several extensions of the discrete Shapley value to continuous settings. Relatedly, this technique is applicable only when the gradient of the prediction score with respect to the base features is well-defined, and is therefore not applicable to models like tree ensembles. 
\end{enumerate}

The first and second approaches solve a different problem (of feature importance across all the training data), and we will ignore them for the most part. Notice that the rest are solving the same attribution problem, and are reflective of the non-uniqueness of the Shapley value for model explanation. \cite{Lundberg2017AUA}  several of these methods (excluding Integrated Gradients) under a common framework based on certain conditional expectations over feature distributions.  However, as we point out later in this paper, the choice of feature distribution influences the attributions significantly, not just in quantity, but also in quality.

\section{Preliminaries}
\label{sec:prelim}
We model the machine-learning model as a real-valued function $f$ that takes a vector of real-valued features as input. If the problem is a classification problem, the function models the \emph{score} of a class. The set of features is denoted by $N$. We designate the input to be explained, i.e., the \emph{explicand}, by the vector $x$ of features; when we say $x_S$ we mean the sub-vector of a vector $x$ restricted to the features in the set $S$.

At times, we may assume that the features are generated according to a distribution $D$; this distribution could be a posited distribution as in Section~\ref{sec:bstoces}. Often, it is the empirical distribution of the training data, wherein it is written as $\hat{D}$. Of special significance are independent feature distributions. The product of marginals of distribution $D$ is written as $\Pi(D)$. The conditional expectation $E[f(x)| x_S]$ is the expected value of the function over the distribution with the features in $S$ fixed at the explicand's value.

\subsection{Shapley value}
\label{sec:shapley}

The Shapley value takes as input a set function $v: 2^N \rightarrow R$. The Shapley value produces attributions  $s_i$ for each player $i\in N$ that add up to $v(N)$. The Shapley value of a player $i$ is given by:

\begin{equation}
\label{eq:shapley}
s_i = \sum_{S \subseteq N \setminus i} \frac{|S|!*(|N| - |S| -1)!}{N!} \left(v(S \cup i) - v(S)\right)
\end{equation}

There is an alternate permutation-based description of the Shapley value: Order the players uniformly at random, add them one at a time in this order, and assign to each player $i$ its expected marginal contribution $V(S \cup i) - v(S)$; here $S$ is the set of players that precede $i$ in the ordering.

In this paper, we study three extensions of the Shapley value to model explanation.

\subsection{Conditional Expectations Shapley (CES)}     
\label{sec:conditional-expecations}
This approach takes three inputs: an explicand $x$, a function $f$, and a distribution $D$. The set function is defined by the conditional expectation
\begin{equation}
v(S) = E_{D}[f(x')|x'_S = x_S)]
\end{equation}
We denote the CES attribution for feature $i$ with explicand $x$, distribution $D$ and function $f$ by $ces_i(x, D, f)$. This approach has been used by~\cite{Strumbelj, Lundberg2017AUA, Datta} and was proposed in this specific form by~\cite{Lundberg2017AUA}, where it is called Shapley Additive Explanations, or SHAP. When CES is carried out with the empirical distribution of the training $\hat{D}$, it will be denoted as CES($\hat{D}$).

\subsection{Baseline Shapley (BShap)}
\label{sec:baseline}
This approach takes as input an explicand $x$, the function $f$ and an auxiliary input called the baseline $x'$. The set function is defined as:
\begin{equation}
v(S) = f(x_S;x'_{N \setminus S})\label{baseline-function}
\end{equation}

That is, we model a feature's absence using its value in the baseline. We call this the Baseline Shapley (BShap) approach. We denote BShap attribution by $bs_i(x, x', f)$. Variants of this approach have been used by ~\cite{Multilinear,  ADS, Lundberg2017AUA}. 

\subsection{Random Baseline Shapley (RBShap)}
This approach is a variant of BShap that takes three inputs: An explicand $x$, a function $f$, and a distribution $D$. The attributions are the expected BShap values, where the baseline $x'$ is drawn randomly according to the distribution $D$. This approach is implicit in~\cite{Lundberg2017AUA} (see Equation 11).
\begin{equation}
v(S) = E_{x' \sim D}f(x_S;x'_{N \setminus S})\label{baseline-function}
\end{equation}

\subsection{Integrated Gradients (IG)}
\label{sec:IG}
This approach takes as input an explicand $x$, the function $f$ and an auxiliary input called the baseline $x'$. 
We consider the straight-line path (in $\reals^{|N|}$) from the baseline $x'$ to the input
$x$, and compute the gradients at all points along the path. The path can be parameterized as
$\gamma(x,\alpha) = x' + \alpha \cdot (x- x')$.
Integrated gradients are obtained by accumulating these gradients. The integrated gradients attribution for an explicand $x$ and baseline
$x'$, for a variable $x_i$ is:
\begin{equation}
\label{eq:IG}
IG_i(x, x',f) = (x_i-x_i')\int_{\sparam=0}^{1} \frac{\partial f(x' + \alpha (x - x'))}{\partial x_i}~d\alpha
\end{equation}
IG is an analog of the Aumann-Shapley method from cost-sharing~\cite{AS74}. We will discuss the sense in which IG is an extension of the Shapley value in Section~\ref{sec:IG-BShap}.

\subsection{Axioms}
\label{sec:axioms}

We now list several desirable properties of an attribution technique and discuss why each property is desirable. Later, we will use these properties as a framework to compare and contrast various attribution methods. Variants of these axioms have appeared in prior cost-sharing literature (cf. ~\cite{FM}).

An attribution method satisfies:
\begin{itemize}
\item\textbf{Dummy} if dummy features get zero attributions. A feature $i$ is dummy in a function $f$ if for any two values $x_i$ and $x'_i$ and every value $x_{N \setminus i}$ of the other features, $f(x_i;x_{N \setminus i}) = f(x'_i;x_{N \setminus i})$; this is just a formal way of saying that the feature is not referenced by the model, and it is natural to require such variables to get zero attributions.

\item \textbf{Efficiency} if for every explicand $x$, and baseline $x'$, the attributions add up to the difference $f(x) - f(x')$ for the baseline approach. For the conditional expectation approach, $f(x')$ is replaced by $E_{x' \sim D}[f(x')]$. This axiom can be seen as part of the framing of the attribution problem; we would like to apportion blame of the entire difference $f(x) - f(x')$ to the features. 

\item \textbf{Linearity} if, feature by feature, the attributions of the linear combination of two functions $f_1$ and $f_2$ is the linear combination of the attributions for each of the two functions. Attributions represent a kind of forced linearization of the function. It is therefore desirable to preserve the existing linear structure in the function.

\item \textbf{Symmetry} if for every function $f$ that is symmetric in two variables $i$ and $j$, if the explicand $x$ and baseline $x'$ are such that $x_i = x_j$ and $x'_i = x'_j$, then the attributions for $i$ and $j$ should be equal. This is a natural requirement with obvious justification. 

\item \textbf{Affine Scale Invariance (ASI)} if the attributions are invariant under a simultaneous affine transformation of the function and the features. That is, for any $c, d$, if $f_1(x_1, \ldots, x_n) =
f_2(x_1, . . . ,(x_i - d)/c, . . . , x_n)$, then for all $i$ we have $attr_i(x, x', f_1) = attr_i((x_1, \ldots , c*x_i + d, \dots x_n),(x'_1, \ldots , c*x'_i + d, \dots x'_n), f_2)$. ASI conveys the idea that the zero point and the units of a feature should not determine its attribution. Here is a concrete example: Imagine a model that takes temperature as a feature. ASI dictates that whether temperature is measured in Celsius or Farenheit, the attribution to the feature should be identical.

\item \textbf{Demand Monotonicity} if for every feature $i$, and function $f$ that is non-decreasing in $i$, the attribution of feature $i$ should only increase if the value of feature $i$ increases, with all else held fixed.
This simply means that if the function is monotone in a feature, that feature's attribution should only increase if the explicand's value for that feature increases.

\item \textbf{Proportionality} If the function $f$ can be rewritten as a function of $\sum_i x_i$, and the baseline (x') is zero, then the attributions are proportional to the explicand values(x). 

\end{itemize}

\subsection{An Empirical Case Study: Diabetes Prediction}

While the bulk of this paper is axiomatic and theoretical, we will replicate some of our observations on a diabetes prediction task. The motivation is to show that many of the issues we identify theoretically show up in practice, and that too in the simplest possible setting, indicating that the issues are commonplace. We train our diabetes prediction models on a data set from the Scikit learning library~\cite{sklearn}; this data set was originally used in~\cite{Efron}. The data has ten base features, age, sex, body mass index (BMI), average blood pressure (BP), and six blood serum measurements. Data is obtained for each of 442 diabetes patients, as well as the response of interest, a quantitative measure of disease progression one year after the time of measurement of the base features. 

We train a linear model using Scikit's implementation of Lasso regression~\cite{Lasso}; we used the standard settings of the fitting algorithm and 75\%-25\% train-test split. The variance explained by the model is 35\%. The model coefficients are ~399 for BMI, ~4.9 for BP and ~291 for the fifth blood serum measurement (s5). The intercept is 154.15, which closely matches the data set average of response. 

%Second, a random forest model with two trees, each of depth two using the same test-train split as we used in the linear model above. The variance explained by this model is 38\%, which is in the same ballpark as the the linear model.

\section{An Analysis of CES}
\label{sec:ces}

While CES was proposed by ~\cite{Lundberg2017AUA}, and justified axiomatically (by citing the original Shapley axiomatization), the justification did not cover the choice of using conditional expectations as the set function (recall the definition of CES in Section~\ref{sec:conditional-expecations}). Furthermore, it appears that CES has only been applied with modification: For instance, ~\cite{Strumbelj}, assumes an independent feature distribution while ~\cite{Lundberg2017AUA} applies it to modules of a deep network rather than end-to-end. Consequently, the properties have CES have not been carefully studied. This is what we remedy in this section. 

\subsection{Comprehending CES($\hat{D}$)}
\label{sec:compute}

As discussed in Sections~\ref{sec:intro} and~\ref{sec:conditional-expecations}, CES attributions depend crucially on the choice of the distribution $D$. Arguably, the most obvious choice is to use the training data distribution $\hat{D}$; we call this CES($\hat{D}$). 

Unfortunately, the properties of CES($\hat{D}$) are not immediately apparent from its definition; the functional forms of the Shapley value and conditional expectations are sufficiently complex as to prevent direct understanding. We therefore begin by redefining CES($\hat{D}$) as an intuitive procedure. 

The input is the (training) data, i.e., a list of examples $T = \{x^{t}\}$.
(We use superscripts to index examples and subscripts to index features.) Given an explicand $x$, $T_S$ is the subset of $T$ that agrees with $x$ on the features in the set $S$, i.e., $T_S = \{x^t| \forall i \in S, x^t_i = x_i\}$. Notice that $T_{\{\}} = T$, and $T_N = \{x_i\}$. The value of the set function $v(S)$ is the average value of the function over inputs in the set $T_S$; this corresponds to computing the conditional expectation  $E[f(x)|x_S)]$ in the CES approach (Section~\ref{sec:conditional-expecations}).

We now use the procedural definition of the Shapley value (see Section~\ref{sec:shapley}), i.e., we average the marginal contribution of 'adding' variable $i$ (i.e., conditioning on it) over permutations of the variables. We notice that conditioning on an additional variable $i$ only reduces examples that 'agree' with $x$ on the conditioned features. We call this the Downward Closure property\footnote{We borrow the term from the frequent itemset mining literature (cf. ~\cite{Agrawal}). In itemset mining, the downward closure property reflects that every subset of a frequent itemset is also frequent. Analogously, for every feature set $S$, every row in $T_S$ is also in 
$T_{S'}$ for every subset $S' \subseteq S$.
}:

\begin{lemma}[Downward Closure]
\label{lem:downward-closure}
For every pair of sets of features $S,S'$ , if $S \subseteq S'$ then $T_{S'} \subseteq T_S$. (Proof in Appendix.)
\end{lemma}

Putting these observations together, we have Algorithm~\ref{ces-alg}. (If needed, the computation can be further sped up by sampling over permutations as is common with the Shapley value, and by caching values of the sets $T_i$ for all $i$.)

\begin{algorithm}
\caption{Computing CES($\hat{D}$)}
\begin{algorithmic}
\STATE Inputs: explicand $x$ and examples $T$, each over feature set $N$
\STATE \COMMENT{Compute Shapley values via permutations}
\STATE $s_{\sigma_i} \leftarrow 0$ for all $i$
\FORALL{permutations $\sigma$ of $N$}
\STATE $v_{new} \leftarrow \frac{1}{|T|}\sum_{x \in T}f(x)$
\STATE $T' \leftarrow T$
\FORALL{$i \in 1 \ldots |N|$}
\STATE $v_{old} \leftarrow v_{new}$
\STATE \COMMENT{Use the Downward Closure Lemma to update $T'$}
\FORALL{$t \in T'$}
\STATE \COMMENT{$\sigma_i$ is the $i$th feature in the ordering $\sigma$}
\IF{$x^t_{\sigma_i} \neq x_i$} 
 \STATE delete $t$ from $T'$
\ENDIF
\ENDFOR
\STATE $v_{new} \leftarrow \frac{1}{|T'|}\sum_{x \in T'}f(x)$
\STATE \COMMENT{Update Shapley value of  $i$th feature in ordering $\sigma$}
\STATE $s_{\sigma_i} \leftarrow s_{\sigma_i} + \frac{1}{|N|!}(v_{new} - v_{old})$
\ENDFOR
\ENDFOR
\end{algorithmic}
\label{ces-alg}
\end{algorithm}

\subsection{The Effect of Sparsity on CES($\hat{D}$)}
Our main motivation for providing a procedure for CES($\hat{D}$) is to understand its properties. Our first observation is that CES($\hat{D}$) is extremely sensitive to the degree of sparsity; sparsity arises naturally when the variables are continuous because it unlikely that data points share feature values precisely.  

\begin{remark}
\label{ex:sparsity}
Suppose we have an explicand $x$ such that every feature value $x_i$ is unique, i.e., it does not occur elsewhere in the training data. Then, notice that $T_S = \{x\}$ for all non-empty sets $S$. Therefore in each permutation, the first feature gets attribution $f(x) - E_{x' ~ D}[f(x')]$ while all the other features get an attribution of zero. Therefore all the variables get \emph{equal} attributions, even if the function is not symmetric in the variables! 
\end{remark}

A practical implication of Example~\ref{ex:sparsity} is that the attributions would be very sensitive to noise in the data. For instance, Figure~\ref{fig:three-methods} shows the distribution of attributions across $20$ explicands for CES($\hat{D}$) (the second column in each plot) on the linear model. The attributions vary across features---for instance BMI has a larger variation than Sex. If we add a tiny amount of noise, and recompute attributions, then \emph{all} the features (including BMI and Sex) will get \emph{identical} attributions (we don't show this in the figure).   

One way to deal with this sensitivity is to \emph{smooth} the data. We can simulate smoothing within Algorithm~\ref{ces-alg}. When we condition on a set $S$ of features in the computation of CES, we average the prediction over all the training data points that are \emph{close} to the explicand in each of the features in $S$; two data points are close in a certain feature if their difference is within a certain fraction of the standard deviation. In our experiments, we use two settings $0.1$ and $0.2$. Figure~\ref{fig:three-methods} shows how different amounts of smoothing change the attributions (see for instance the attributions of the feature S2). Thus while smoothing mitigates sensitivity, it is still unclear how much smoothing to do. 

There are some other approaches to dealing with sparsity. One approach is to use the distribution $\Pi(\hat{D})$, i.e., the product of empirical marginal distributions, as in~\cite{Datta}, or to assume that the function is somewhat smooth, and to compute the function's value at a point outside the training data using a weighted sum of nearby points in the training data as in~\cite{Aas}. Again, these will undoubtedly give different results, and we cannot easily pick between them.

\subsection{An Axiomatic Analysis of CES($\hat{D}$)}
\label{sec:distortion}

As discussed in Section~\ref{sec:shapley}, the Shapley value and its variants were conceived in the context of cooperative game-theory where there is no analog of the feature distribution $D$. The axioms (see Section~\ref{sec:axioms}) were meant to guarantee that if the set function has certain properties (the antecedent), then the Shapley values must have certain properties (the consequent). For instance, the Dummy axiom says that if a function is insensitive to a feature (the antecedent), then the feature should have zero attributions (the consequent). However, there are two functions at work here: the function $f$ whose value we wish to attribute, and the set function $v$ used to compute Shapley values.

CES defines $v(S)$ using conditional expectations that depend both on the function $f$ and the distribution $D$. Consequently, even if the function $f$ satisfies certain properties, the consequent property of the axiom need not hold for $v$. This gives rise to counter-intuitive attributions. We give several such examples in this section. These do not imply that prior axiomatizations (cf. ~\cite{Lundberg2017AUA} or~\cite{Datta}) are incorrect. The various axioms still hold for the $v$ based on the conditional expectation (see Section~\ref{sec:conditional-expecations}). However, the axioms have no natural interpretation for this set function. In contrast, it is natural to seek interpret axioms as properties of the model function $f$.

In all our examples, each row of the table is a combination of feature values; the first column specifies the probability of this feature combination, the next two columns specify the feature values for two discrete, abstract features $T$ and $B$, and the remaining columns specify the value of the function for this feature combination. Feature and function values used as explicand in the examples are in bold.

\begin{example} [Failure of Dummy]
See the function $f_1$ in Table~\ref{tab:dummy} modeled as a bivariate function of $x$ and $y$. For the explicand $x=5, y=5$,  the CES attributions are $\frac{22.5}{2}$ each for $x$ and $y$ (a consequence of Remark~\ref{ex:sparsity}). Therefore  the variable $x$ gets a large attribution despite being dummy. 
\end{example}

\begin{table}[!htb]
    \centering
    \small
    \begin{tabular}{llllll}
    \hline
     Probability& $x$ & $y$ & $f_1$=$y^2$ & $f_2 =x$ & $f_1 + f_2$\\
    \hline
    $\epsilon$ & \textbf{5} & \textbf{5} &\textbf{25} &\textbf{5} &\textbf{30}\\
    $\frac{1-\epsilon}{2}$& 1 & 1 &1 &1 &2\\
    $\frac{1-\epsilon}{2}$& 1 & 2 &4 &1 &5 \\
    \hline
    \end{tabular}
    \caption{Example for: (a) Dummy, correlated variables can have large CES attributions. (b) CES attributions are not linear in the function.}
    \label{tab:dummy}
\vspace{-4mm} %reduce too much white space
\end{table}

\begin{figure*}
  \centering
      \includegraphics[scale=0.7]{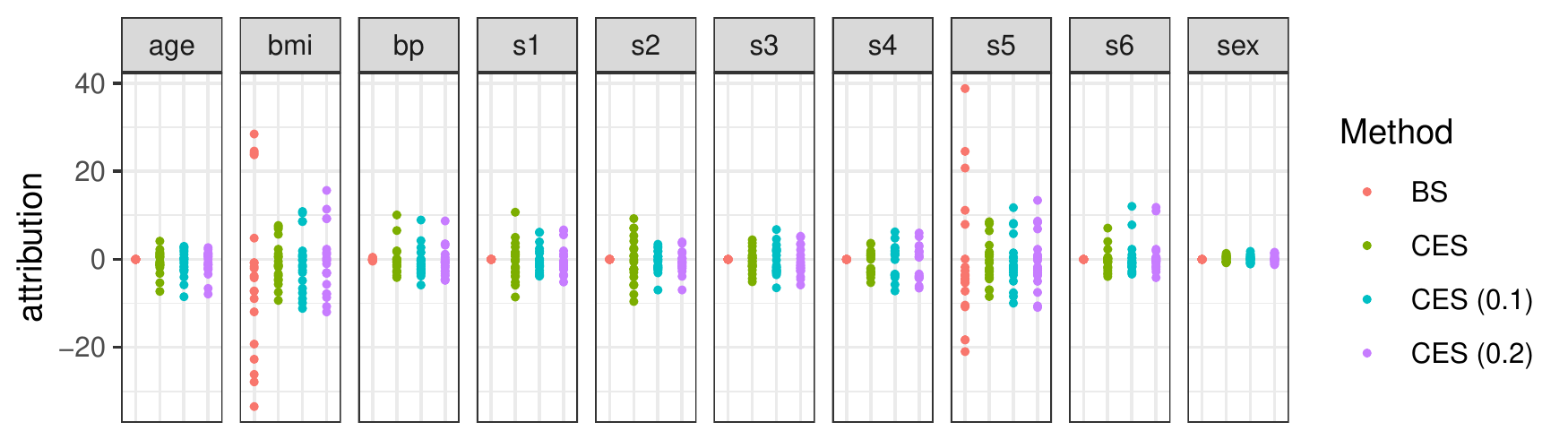}
  \caption{Attribution distribution across 20 explicands for four methods, BShap, CES, CES (smoothing $0.1$),  CES (smoothing $0.2$).}
  \label{fig:three-methods}
\end{figure*}

The implication is this: Say, in the context of an analysis of fairness, we require that a certain feature play no role in the prediction model, and indeed, it does not. If we use CES, it may still be assigned significant attribution, leading us to incorrectly believe that the function is sensitive to the variable. In our diabetes prediction task, for the linear model (see Figure~\ref{fig:three-methods}), we note that 7 of the 10 variables are dummy features. Despite this, CES assigns non-zero attributions to them. (In contrast, BShap assigns zero attributions to the dummy features.)

\begin{example} [Failure of Linearity]
 See the functions $f_1$ and $f_2$ in Table~\ref{tab:dummy}; model them as univariate functions of $y$ and $x$ respectively. Consider the explicand $x=5,y=5$. Then the CES for the variable $y$ with the function $f_1$ is the difference between the function value at the explicand ($25$) and the mean of the function ($2.5$), i.e., $22.5$, and that for the function $f_2$ is zero ($y$ does not appear in the function). Now consider the attribution of $y$ for the function $f_1 + f_2$; both variables get an attribution of $\frac{30-3.5}{2}$ (again, a consequence of Remark~\ref{ex:sparsity}), which is not $22.5 +0$.
\end{example}

Here is an implication: Imagine, if we were computing attributions for an ensemble of trees. Recall that the prediction of the forest is a uniform average over the trees, i.e., it is linear in the prediction of the trees.  Therefore, we would expect the attributions to also be linear. But this is not the case. We saw large failures in linearity for the diabetes prediction task even for a two tree ensemble with trees of depth two. 

See Appendix~\ref{app:axioms} for how CES($\hat{D}$) fails the Symmetry and Demand Monotonicity.

\section{Baseline Shapley and its Properties}

 In this section, we discuss the properties of BShap and provide a proper axiomatic result for it. (Recall the definition of Baseline Shapley (BShap) from Section~\ref{sec:baseline}.) As discussed in the introduction, prior axiomatization results from the machine learning literature (e.g. \cite{Datta, Strumbelj,Lundberg2017AUA}) did not cover the choice of function input to the Shapley value, and consequently there are a multiplicity of methods that yield different results.

\subsection{BShap versus IG}
\label{sec:IG-BShap}
The model explanation literature has largely built on top of the Shapley value from the binary cost-sharing literature. In this literature, players (features) are either present or absent. In contrast, machine learning tends to involve continuous features and deep learning involves only continuous features--even discrete/categorical are often turned into continuous features via embeddings. Therefore, it is worth connecting model explanation with the rich continuous cost-sharing literature. 

 We begin by defining cost-sharing formally. 
 
 A \emph{cost-sharing problem} is an attribution problem with function $f$, explicand $x$ and baseline $x'$ such that the baseline $x'=0$, the explicand $x$ is non-negative, and the function $f$ is non-decreasing in each variable, i.e., if two feature vectors $x^- \leq x^+$ (point-wise for every feature), then $f(x^-) \leq f(x^+)$.\footnote{In~\cite{FM}, the function is defined to satisfy an additional property of being zero at $f(0)$; but this is only used to simplify the definition of the efficiency axiom (see Section~\ref{sec:axioms} to not require the $f(0)$ term.}

There are several extensions of the Shapley value to the continuous cost sharing literature (see~\cite{FM} for details). Two of these extensions correspond to BShap and IG.  BShap is a generalization of a classic cost-sharing method called Shapley-Shubik (cf. ~\cite{FM}), and IG is a generalization of a cost-sharing method called Aumann-Shapley (cf. ~\cite{AS74}). One can get Shapley-Shubik from BShap and Aumann-Shapley from IG by setting the baseline $x'$ to zero; the explicand value $x_i$ corresponds to the demand of player $i$, the function $f$ corresponds to the cost incurred, and the attributions to the cost-shares. 

It is relatively clear how Shapley-Shubik (BShap) is an extension of the binary Shapley value from its definition (see Section~\ref{sec:baseline}). 

However it is less clear how Aumann-Shapley (IG) (Equation~\ref{eq:IG}) is an extension of the binary Shapley value. IG traverses a single, smooth path between the baseline and the explicand, and aggregates the gradients along this path. Whereas the Shapley value takes an average over several discrete paths---in each step of a discrete path, a variable goes from being 'off' to 'on' in one shot. 
To establish the connection, notice that the IG path can be seen to be the internal diagonal of a $N$ dimensional hypercube, and in contrast, the Shapley value is an average over the extremal paths over the edges of this hypercube.  Suppose we partition every feature $i$ into $m$ micro features, where each micro feature represents a discrete change of the feature value of $\frac{x_i - x'_i}{m}$. And then we apply the Shapley value on these $N*m$ features. Notice that this is equivalent to creating a grid within the hypercube, and averaging over random, monotone walks from the baseline $x'$ to the explicand $x$ in this grid. As $m$ increases, the density of the random walks converges to the diagonal of the hypercube, and if the function $f$ is smooth, then running Shapley on these micro-features is equivalent to running IG on the original features.

Of course, in general, IG and BShap use different paths and hence give different attributions (see Example~\ref{ex:min}). 

The standard axiomatization of the Shapley value \cite{Shapley53} only references (binary variants of) the first four axioms (Dummy, Efficiency, Linearity and Symmetry). However, in the continuous setting there are infinitely many methods that satisfy these four axioms. A uniqueness result requires further axioms, as we study in the next section.

\subsection{Axiomatizing BShap (and IG)}
\label{sec:axiomitization}

In this section, we provide axiomatizations for BShap and IG by formally reducing model explanation to cost-sharing:

\begin{theorem}[Reducing Model Explanation to Cost-Sharing]
\label{thm:reduction}
Suppose there is an attribution method that satisfies Linearity and ASI. Then for every attribution problem with explicand $x$, baseline $x'$ and function $f$ (satisfying the minor technical condition that the derivatives are bounded) , then there exist two cost-sharing problems such that the resulting attributions for the attribution problem are the  difference between cost-shares for the cost-sharing problems.
\end{theorem}
\begin{proof}
Given an attribution problem $f,x,x'$, we progressively transform it into equivalent cost-sharing problems. 

First, we transform the problem $f,x',x$ into a problem $f^n,x'^n, x^n$ such that the baseline $x'^n$ is the zero vector,
and $x^n$ is non-negative. The proof is inductive. The base case is by definition: we define $f^0,x^0, x'^0$ to be $f,x, x'$. In step $i$ we transform $f^{i-1},x^{i-1}, x'^{i-1}$
into $f^{i},x^{i}, x'^{i}$ by transforming along feature $i$ using the transformation in the definition of ASI (see Section~\ref{sec:axioms}) using the $c=1$ if $x_i$ is non-negative and $c=-1$ otherwise, and $d=-x'_i*c$. The proof for the inductive step: By ASI, the attribution for $f^{i-1},x^{i-1}, x'^{i-1}$ should equal to that for $f^{i},x^{i}, x'^{i}$, and we have set the baseline value for this feature to $0$, and its explicand value is non-negative. 

Next we express the function $f^n$, as the difference of two non-decreasing functions $f_1$ and $f_2$. Let $p$ denote the infimum of the partial derivative $\frac{\partial f^n(x)}{\partial x_i}$, where $x$ ranges over the domain of the function $f^n$, and $i$ ranges over all the variables. By the technical conditions, this infimum exists. If $p$ is zero or positive, then $f^n$ is itself non-decreasing---therefore set $f_1$ to $f^n$ and $f_2$ to the constant zero function. Otherwise, define $f_2$ to be the linear function $\sum_i -p * x_i$; notice that $-p$ is positive and so the function is non-decreasing. Set $f_1 = f^n + f_2$; by definition of $p$, $f_1$ is non-decreasing. By Linearity, the attributions for $f^n,x^n, x'^n$, (which we have already shown are equal to the attributions for $f,x, x'$) is the difference between the attributions of $f_1,x^n, x'^n$ and $f_2,x^n, x'^n$. 

To complete the proof, notice that both  $f_1,x^n, x'^n$ and $f_2, x^n, x'^n$ are cost-sharing problems. 
\end{proof} 

Both IG and BShap satisfy ASI and Linearity. Therefore \emph{any} axiomatization that applies to Aumann-Shapley applies to IG and any axiomatization that applies to Shapley-Shubik applies to BShap. For instance, Corollary 1 from~\cite{FM} reads: 

\begin{theorem}
\label{thm:cost-sharing}
Shapley-Shubik is the unique method that satisfies the Efficiency, Linearity, Dummy, Affine Scale Invariance (ASI), Demand Monotonicity (DM) and Symmetry (plus some technical conditions we exclude for clarity) for all cost-sharing problems.
\end{theorem}

Therefore we have:

\begin{corollary}
\label{cor:main}
 BShap is the unique method that satisfies the Linearity, Dummy, Affine Scale Invariance (ASI), Demand Monotonicity (DM), and Symmetry (plus minor technical conditions) for all attribution problems. (See Proof in Appendix.)
\end{corollary}

Analogously, we have the following corollary of Theorem 3 from~\cite{FM}: 

\begin{corollary}
\label{cor:IG}
 IG is the unique method that satisfies the Linearity, Dummy, Affine Scale Invariance (ASI), Proportionality, and Symmetry (plus minor technical conditions) for all attribution problems. 
\end{corollary}

\begin{remark} [BShap versus IG]
\label{ex:min}
A simple example where the IG and BShap differ is the min of two variables $x_1$ and $x_2$. Suppose the baseline is $x'_1=x'_2=0$, and the explicand is $x_1=5, x_2=1$. IG attributes the entire change in the function of $1$ to the arg min, i.e., $x_2$; this is intuitively reasonable if you see $x_2$ as the critical variable. BShap on the other hand assigns attributions of $2.5$ to the first variable and $-1.5$ to the second; this is intuitively reasonable if you think of $x_1$ as trying to increase the min and $x_2$ as trying to decrease it. It is not immediately clear which interpretation is obviously superior.

Let us now consider the cube of the sum of two variables $x_1$ and $x_2$. Suppose the baseline is $x'_1=x'_2=0$, and the explicand is $x_1=5, x_2=1$. IG attributes $180$ to $x_1$ and $36$ to $x_2$, i.e., the attributions are in the ratio $5:1$, a consequence of the proportionality axiom. BShap attributes $170$ to $x_1$ and $46$ to $x_2$. The IG results appear a bit more principled. But there is a stronger consequence of the proportionality axiom: This axiom forces a smooth interpolation between the baseline and the explicand, i.e, the form of IG. For instance, a baseline of an entirely black image used in computer vision tasks results in intermediate inputs that are variants of the explicand with different intensities. In contrast, BShap is likely to construct more unrealistic inputs; in the vision examples, mixes of black pixels with the explicand pixels. 
\end{remark}

\subsection{BShap fits in the CES framework}
\label{sec:bstoces}

Thus far, we have studied the source of the difference between IG and BShap, both in terms of computation (they take different paths through a feature grid) and in terms of axioms (Corollary~\ref{cor:main} versus Corollary~\ref{cor:IG}). We now study the difference between BShap and CES. 

First, we show that one can \emph{posit} a distribution $D$ such that the BShap approach coincides with CES under the distribution $D$\footnote{\cite{Lundberg2017AUA} shows that CES reduces to BShap if the baseline is the feature means, \emph{if} the features are independently distributed \emph{and} the model is linear. Our reduction applies to non-linear models and baselines other than the feature means.}. This shows that BShap fits in the framework of CES. In this sense, any difference in the properties of BS and CES($\hat{D}$) can be isolated to the choice of distribution on which to run CES.

\begin{lemma}
\label{lem:bshap_is_ces}
For any explicand $x$ and baseline $x'$, there exists a feature distribution such that CES over that distribution results in attributions that are arbitrarily close to those produced by BShap. (Proof in Appendix)
\end{lemma}

\begin{remark}[Explicit Comparison]
\label{rem:explicit}
Unlike CES($\hat{D}$), BShap does not depend on any distribution but requires an additional input (the baseline). We can use the baseline to model the explanation context. For instance, consider a model that makes loan decisions. If the applicant has been denied a loan, it is likely more useful to produce an explanation that only attributes to features that are in the applicant's power to change (e.g. getting a high school diploma as opposed to reducing their age). Such an explanation is achievable by selecting a baseline that coincides with the explicand on immutable features. In this sense the baseline parameter is useful flexibility. It makes the attributions germane to the decision or actions of the person consuming the explanation. However, this does pose an additional cognitive load to select the baseline and interpret the dependence of the explanation on the baseline. There are also situations where there is no compelling choice of one baseline over another. In this situation, RBShap presents an alternative. 
\end{remark}

\subsection{CES and RBShap}

In Section~\ref{sec:ces}, we saw that CES violates axioms like Dummy and Linearity. In contrast, Theorem~\ref{cor:main} shows that BS satisfies these axioms. But since Lemma~\ref{lem:bshap_is_ces} shows that BS is a variant of CES, one could ask if there are other variants of CES that also satisfy some of the axioms.

We first note that several axioms (Dummy, Linearity, Demand Monotonicity) also apply to RBShap, because these are satisfied by BShap and preserved by averaging the attributions over several baselines. (Symmetry requires that the distribution $D$ is symmetric in features over which the function $f$ is symmetric.)
We now note that CES over an independent distribution $D$ is equal to RBShap where the baselines are drawn from distribution $D$. Therefore, these axioms also carry over to CES \emph{if} the feature distribution $D$ is independent. 

\begin{lemma}
\label{lem:rbs_to_ces}
If the distribution $D$ is an independent distribution over the features, then RBShap and CES coincide. (Proof in Appendix)
\end{lemma}

\begin{remark}
\label{rem:linear}
If the function $f$ is linear and the distribution is independent, then, BShap with the baseline vector that has each feature set to its average across the data set, has the same attributions as CES (a consequence of Equation 9-12 from~\cite{Lundberg2017AUA}) and hence RBShap (due to Lemma~\ref{lem:rbs_to_ces}). 
\end{remark}

\begin{remark}
\label{rem:marginal}
\cite{Datta} runs CES over an independent feature distribution $D'$ that is a product of the marginal distributions of the input feature distribution $D$. By Lemma~\ref{lem:rbs_to_ces}, CES over $D'$ is equivalent to RBShap over $D'$. One could ask if RBShap on the contrived feature distribution $D'$ is equivalent to RBShap on the original feature distribution $D$. The following example proves that this is false, by showing that the sum of the attributions in the two cases differ: Suppose we are given two binary features $x_1$ and $x_2$, function $f(x_1, x_2) = x_1 * x_2$ and a distribution $D$ such that $P(0, 0)= P(1, 1) = 0.5$. Suppose that the explicand is $x_1=1, x_2=1$. Under $D$, $E[f(x)] = 0.5$. 
Under $D'$, $E[f(x)] = 0.25$. Recall that RBShap attributions sum to $f(x) - E[f(x)]$; this completes our counterexample. This shows that forcing independence by constructing a marginal distribution is different from working with an independent distribution.
\end{remark}

\begin{remark}
Though CES over an independent distribution satisfies several axioms, it can fail Strong Monotonicity as the example in Table~\ref{tab:monotonicity} shows; the distribution in this example is independent.
\end{remark}

\begin{remark}
The examples in Section~\ref{sec:distortion}, show that if the distribution is not independent, CES can fail Linearity, Dummy and Demand Monotonicity. However, it always satisfies Affine Scale Invariance (we omit the easy but technical proof). 
\end{remark}

\section{Conclusions}
We show that Shapley with Conditional Expectations is highly sensitive to data sparsity, and can produce counterintuitive attributions. We produce proper axiomatizations (uniqueness results) for Baseline Shapley and Integrated Gradients.

\bibliography{main}
\bibliographystyle{acm}

\newpage

\appendix

\section{Strong Monotonicity}
We introdoce an additional axiom called Strong Monotonicity that plays a role in the results in the Appendix.

\textbf{Strong Monotonicity} if for every two functions, $f_1$ and $f_2$, on the same domain, if for some feature $i$, at all points $x$ in the domain, $\frac{\partial f_2(x)}{\partial x_i} \geq \frac{\partial f_1(x)}{\partial x_i} \geq 0$, then the magnitude of attribution of feature $i$ for function $f_2$ is at least as large as that for function $f_1$\footnote{A slight generalization of the cost-sharing axiom from \cite{Young}.}. A simple example will help clarify this axiom: Imagine that $f_1$ and $f_2$ are both linear models on an identical feature set. Moreover, the coefficients of all the features except one (call this feature $i$) are identical; say that $f_2$ has a large positive coefficient for this feature $i$, while $f_1$ has a small positive coefficient. This satisfies the conditions on the partial derivative in the antecedent of the axiom. 
Since the coefficient of feature $i$ in function $f_2$ is larger, it is reasonable for feature $i$ to have an attribution of greater magnitude for $f_2$ than for $f_1$. 

\section{Issues with prior Axiomatic Results}

\cite{Lundberg2017AUA} claims to show that one of the standard axioms (Symmetry) used in the Shapley axiomatization is redundant within an earlier axiomatization of the Shapley value by~\cite{Young}. It claims that Missingness, Local Accuracy and Consistency suffice. (The latter two axioms are called Efficiency and Strong Monotonicity by~\cite{Young}). This claim is incorrect. Here is the counterexample: Suppose that the function is $x_1 * x_2 * x_3$, with all three feature values identically $1$ in the explicand. Assume that the features are independently and identically distributed over the discrete set $\{0,1\}$, with a probabilities $1-\epsilon$ and $\epsilon$ for values $0$ and $1$ respectively. Shapley gives identical shares to all three variables, $1/3$. Now, consider an alternate attribution method: Take a fixed permutation of the three variables, $x_1 \rightarrow x_2 \rightarrow x_3$ and define the attributions to be the marginals of this permutation as in Section~\ref{sec:shapley}. This method satisfies missingness, local accuracy, and consistency, but yields different attributions from Shapley values---the third variable gets an attribution of $~1$ and the rest get attributions of $0$.

The axiomatization for Deep Shap in \cite{Lundberg2017AUA} is problematic in a different way: In Deep Shap, the 
Shapley value is applied layer by layer rather than to the network as a whole. This destroys the guarantees of the Shapley axioms because the attributions become sensitive to the arrangement of parameters in the network, as opposed to the function that the network computes. Here is an an analogy using a simple function $x1*x2*x3$. Consider an `implementation’ of the function first computes $x1*x2$ and then multiplies it with $x3$, i.e., $((x1*x2)*x3)$. Let us say that the feature values are all identically $1$ in the explicand and identically $0$ in the baseline. Deep Shap would attribute half to $x3$ and half to the product $x1*x2$, it would then redistribute the half equally among $x1$ and $x2$, resulting in attributions that are a $1/4$ each for $x1$ and $x2$ and $1/2$ for $x3$. If the ‘implementation’ was $(x1*(x2*x3))$ instead, then the attributions would be $1/2$ for $x1$ and $1/4$ each for $x2$ and $x3$. Notice that both attributions are also a violation of the symmetry axiom. Applying the BShap end-to-end would result in attributions of $\frac{1}{3}$ each.

\section{How CES($\hat{D}$) fails other Axioms}
\label{app:axioms}

\begin{example}[Failure of Demand Monotonicity]
See the example in Table~\ref{tab:demand-monotonicity}. The CES attribution for feature $y$ for the explicand $x=1, y=0$ (a positive number) exceeds that for the explicand $x=1, y=1$ (a negative number); Notice that the explicands differ only the value of feature $y$ and the function $f$ is monotone; therefore this is a failure of Demand Monotonocity. 
\end{example}

\begin{table}[!htb]
    \centering
    \small
    \begin{tabular}{llllll}
    \hline
     Probability& $x$ & $y$ & $f$=$100*x + y$\\
    \hline
    $1/3$ & $\textbf{1}$ & \textbf{1} & \textbf{101}\\
    $1/3$ & \textbf{1} & \textbf{0} &\textbf{100}\\
    $1/3$ & 0 & 1 & 1\\    
    \hline
    \end{tabular}
    \caption{Increasing the feature value of $y$, can reduce its CES attribution, even though $f$ is monotone.}
    \label{tab:demand-monotonicity}
\vspace{-4mm} %reduce too much white space
\end{table}

An implication of the above example is that it may lead the consumer of the explanation to believe that the function is non-monotone, even if this is not the case.

\begin{example} [Failure of Symmetry]
 In Table~\ref{tab:symmetry}, the function is symmetric in both variables. Furthermore, the variables are distributed independently; variable $T$ is $2$ with probability $p$ and $1$ with the remaining probability; variable $B$ is $2$ with probability $q$ and $1$ with the remaining probability. For the symmetric explicand $T=2,B=2$, variable $T$ gets attribution $1-p$ and variable $B$ gets attribution $1-q$, a violation of symmetry when $p \neq q$.
 \end{example}

\begin{table}[!htb]
    \centering
    \small
    \begin{tabular}{llll}
    \hline
     Probability& $x$& $y$ & $f=x+y$\\
    \hline
    $(1-p)*(1-q)$ & 1 & 1 &2\\
    $(1-p)*q$ & 1 & 2 &3\\
    $p*(1-q)$ & 2 & 1 &3\\
    $p*q$ & \textbf{2} & \textbf{2} &\textbf{4}\\ 
    \hline
    \end{tabular}
    \caption{CES can give symmetric variables unequal attributions, even if the distribution $D$ is independent.}
    \label{tab:symmetry}
\vspace{-4mm} %reduce too much white space
\end{table}
 
An implication of the above example is that it may lead the consumer of the explanation to believe the two variables are not symmetric, even if they actually are.
 
\begin{example} [Failure of Strong Monotonicity]
  Table~\ref{tab:monotonicity} shows an example where the function $f_2$ has larger partial derivatives for variable $x$ than $f_1$ at each point in its domain (the domain is $x\geq1, y \geq 1$). Also, the two variables are independently distributed. Now consider the explicand $x=2,y=2$. The attribution to the variable $x$ is lower for $f_2$ than for $f_1$ (going from $\frac{2\sqrt{2} -1 - \sqrt{3}}{3} \approx 0.032$ to $0$), a violation of Strong Monotonicity. 
\end{example}

\begin{table}[!htb]
    \centering
    \small
\begin{tabular}{lllll}
    \hline
     Probability& $x$ & $y$& $f_1 = \sqrt{x} +y$ & $f_2=x + y$\\
    \hline
    $1/6$ & 1  & 1 & 2 &2 \\
    $1/6$ & 2 & 1 & 1 + $\sqrt{2}$ &3  \\
    $1/6$ & 3 & 1 & 1 + $\sqrt{3}$ &4  \\
    $1/6$ & 1  & 2 & 3 &3  \\
    $1/6$ & \textbf{2} & \textbf{2} & $\mathbf{2 + \sqrt{2}}$ &\textbf{4} \\
    $1/6$ & 3 & 2 & 2 + $\sqrt{3}$ &5  \\   
    \hline
    \end{tabular}
    \caption{Increasing a variable's influence can reduce its CES attributions, even when the distribution is independent.}
    \label{tab:monotonicity}
\vspace{-4mm} %reduce too much white space
\end{table} 
  
An implication of this example is that even if we strengthen a variable's influence, its CES attribution could nevertheless fall. This is because the function transformation influences the function value on points in the background distribution more strongly than at the explicand. 

We did not observe significant failures of Demand Monotonicity in this empirical study; both models we studied were asymmetric in the variables, so no failure of Symmetry was possible.

\section{Proofs}

\subsection{Proof of Lemma~\ref{lem:downward-closure}}
\begin{proof}
Consider an $x^t$ that belongs to $T_{S'}$. We note that it also belongs to $T_{S}$. This is because an example that agrees with the explicand on a feature set $S'$ also agrees with the explicand on every feature set $S$ that is a subset of $S'$.
\end{proof}

\subsection{Proof of Lemma~\ref{lem:bshap_is_ces}}
\begin{proof}
We construct the feature distribution for CES as follows: the distribution for feature $i$ has two points in its support, $x_i$ and $x_i'$, where $\Pr(x_i)=\epsilon$ and $\Pr(x'_i)=1- \epsilon$.
It suffices to show that the set functions input to the Shapley value for the two approaches have values that are arbirarily close when $\epsilon \rightarrow 0$. The set function for CES is:

\begin{equation}
v(S) = \frac{\sum_{S' \subseteq N \setminus S} 
f(x_{S' \cup S};x'_{N \setminus (S' \cup S)})
\epsilon^{|S'|} *  (1- \epsilon)^{|N \setminus (S' \cup S)|}}{\sum_{S' \subseteq N \setminus S} \epsilon^{|S'|} *  (1- \epsilon)^{|N \setminus (S' \cup S)|}}
\end{equation}

Because $\epsilon \rightarrow 0$, the numerator is dominated by the term where $S'$ is empty,  and the numerator tends to $f(x_S;x'_{N \setminus S}) * (1-\epsilon)^{|N \setminus S|}$. By an analogous argument, the denominator tends to $(1-\epsilon)^{|N \setminus S|}$. Dividing, we get the set function for BShap.
\end{proof}

\subsection{Proof of Lemma~\ref{lem:rbs_to_ces}}
\begin{proof}
In the special case that the features follow an independent distribution we show that the set function $v(S)$ for RBShap and CES are the same for all sets $S$. Fix a set $S$ and consider $v(S)$ for RBShap:

\begin{eqnarray}
v(S)
&=& E_{x' \sim D} f(x_S;x'_{N \setminus S}) \\
&=& E_{x'_{N \setminus S}} f(x_S;x'_{N \setminus S}) \label{dummy-conditioning}\\
&=& E_{x'_{N \setminus S}} [f(x_S;x'_{N \setminus S}) | x'_S=x_S] \label{vacuous-conditioning}\\
&=& E[f(x)|x_S]
\end{eqnarray}
where \ref{dummy-conditioning} follows because the expression is dummy in $x'_S$; \ref{vacuous-conditioning} is due to feature independence; and the final expression is the set function for CES.
\end{proof}

\subsection{Proof of Corollary~\ref{cor:main}}
\begin{proof}
It is easy to show that BShap satisfies all the axioms (we skip this part). Now consider the reverse direction, i.e., we would like to show that no other method satisfies these axioms. By Theorem~\ref{thm:reduction}, and because the attribution method satisfies Linearity and ASI, the attributions for an attribution problem are the difference between the cost-shares for two cost-sharing problems, and these cost-shares are uniquely determined by Theorem~\ref{thm:cost-sharing}. 
\end{proof}

\end{document}